# A GP-MOEA/D Approach for Modelling Total Electron Content over Cyprus

Andreas Konstantinidis*, Haris Haralambous*, Alexandros Agapitos† and Harris Papadopoulos* *Department of Computer Science and Engineering, Frederick University, Nicosia, Cyprus
{com.ca, eng.hh, h.papadopoulos}@frederick.ac.cy †School of Computer Science and Informatics, University College, Dublin, Ireland, alexandros.agapitos@ucd.ie

*Abstract*—Vertical Total Electron Content (vTEC) is an ionospheric characteristic used to derive the signal delay imposed by the ionosphere on near-vertical trans-ionospheric links. The major aim of this paper is to design a prediction model based on the main factors that influence the variability of this parameter on a diurnal, seasonal and long-term time-scale. The model should be accurate and general (comprehensive) enough for efficiently approximating the high variations of vTEC. However, good approximation and generalization are conflicting objectives. For this reason a Genetic Programming (GP) with Multi-objective Evolutionary Algorithm based on Decomposition characteristics (GP-MOEA/D) is designed and proposed for modeling vTEC over Cyprus. Experimental results show that the Multi-Objective GP-model, considering real vTEC measurements obtained over a period of 11 years, has produced a good approximation of the modeled parameter and can be implemented as a local model to account for the ionospheric imposed error in positioning. Particulary, the GP-MOEA/D approach performs better than a Single Objective Optimization GP, a GP with Non-dominated Sorting Genetic Algorithm-II (NSGA-II) characteristics and the previously proposed Neural Network-based approach in most cases.

## I. INTRODUCTION

The ionosphere is defined as a region of the earth's upper atmosphere where sufficient ionisation can exist to affect the propagation of radio waves. It ranges in height above the surface of the earth from approximately 50 km to 1000 km. The influence of this region on radio waves is accredited to the presence of free electrons. The impact of the ionosphere on communication, navigation, positioning and surveillance systems is determined by variations in its electron density profile and total electron content along the signal propagation path [1], [2]. As a result satellite systems for communication, navigation, surveillance and control that are based on trans-ionospheric propagation may be affected by complex variations in the ionospheric structure in space and time. This often leads to degradation of accuracy, reliability and availability of their service. Vertical Total Electron Content (vTEC) is an important parameter in trans-ionospheric links since when multiplied by a factor which is a function of the signal frequency, it yields an estimate of the delay imposed on the signal by the ionosphere due to its dispersive nature.

This paper describes an attempt to develop a model to predict vTEC over Cyprus and encapsulate its variability on a diurnal, seasonal and long-term scale. The model development is based on around 60000 hourly vTEC measurements recorded above Cyprus from 1998 to 2009. The practical application of this model lies in its possible use as an alternative candidate local model to the existing Klobuchar global model [3] that is currently being used in single frequency GPS navigation system receivers to improve positioning accuracy.

Metaheuristics and more specifically Evolutionary Algorithms were shown efficient and effective in dealing with difficult-to-solve real-life problems [4]. Particularly, Genetic Programming (GP) based approaches performed well in evolving computer programs, controllers and models [5] in the past. GP approaches deal with this kind of problems by learning from historical data and designing a model for predicting future events. One of the major drawbacks of GP approaches is their bias towards improving their predictive accuracy on the examples available for training [6]. This often results in having a good approximation while evolving the model and a poor approximation in predicting future events, especially in highly distorted cases. In this paper, we have designed a Genetic Programming (GP) approach with a Multi-Objective Evolutionary Algorithm based on Decomposition (MOEA/D) [7] characteristics, coined GP-MOEA/D, for alleviating the aforementioned drawback and dealing with the vTEC prediction problem in the context of Multi-Objective Optimization (MOO) [8]. In MOO, there is no single solution that optimizes all objectives in a single run, but a set of mathematically equally important (or non-dominated) solutions, commonly known as the Pareto Front (PF) [8]. Therefore, our major goal is to obtain a set of Pareto-optimal models, i.e. with high predictive accuracy on the training data as well as comprehensive and general enough.

The main contribution of our paper is as follows:
- A newly proposed vTEC prediction problem is formulated in the context of MOO, using a real data set of vTEC measurements recorded over Cyprus for a period of 11-years.
- A GP-MOEA/D approach, i.e. a panmictic, generational, elitist Genetic Programming (GP) approach having characteristics of the Multi-Objective Evolutionary Algorithm based on Decomposition (MOEA/D) with an expression-tree representation, is designed for dealing with the vTEC prediction problem.
- A GP-based prediction model is derived for vTEC over Cyprus showing a better performance than a Single Objective Optimization GP, a GP with Non-Dominated Sorting Genetic Algorithm-II (NSGA-II) [9] characteris-tics and the previously proposed Neural Network [10] models.

The rest of the paper is organized as follows: Section II introduces background material and related work. Section III defines the vTEC prediction problem by describing vTEC characteristics and measurements performed during the period 1998-2009. The proposed approach is detailed in Section IV. The experimental methodology and results are reported and discussed in Sections V and VI respectively. Section VII concludes the paper.

## II. BACKGROUND AND RELATED WORK

The importance of accurate spatial and temporal vTEC specification [11] in the context of a wide spectrum of space-based telecommunication, radar and navigation systems was a decisive factor encouraging a number of studies with various modeling approaches and prediction techniques [12]. These techniques have ranged from statistical time-series analysis [13] and harmonic analysis [14], [15] to AI techniques. Neural networks were widely adopted as a favourable option in ionospheric modeling [16] and specifically for vTEC, for which local [10] and regional models have been published [17]. Additional studies have also been conducted in the application of relevant techniques in vTEC modelling such as recurrent [18] and radial basis function (RBF) [19] neural networks.

Genetic Programming (GP) [5] is an Evolutionary Computation (EC) technique that evolves populations of computer programs as solutions to problems. The term evolutionary algorithm [4] describes a class of stochastic search processes that operate through a simulated evolution process on a population of solution structures, which represent candidate solutions in the search space. Evolution occurs through (i) a selection mechanism that implements a survival of the fittest strategy, and (ii) diversification of the selected solutions to produce offspring for the next generation. In GP, programs are usually expressed using hierarchical representations taking the form of *syntax-trees*. It is common to evolve programs into a constrained, and often problem-specific user-defined language. The variables and constants in the program are leaves in the tree (collectively named as terminal set), whilst arithmetic operators are internal nodes (collectively named as function set). GP finds out how well a program works by running it, and then comparing its behaviour to some ideal, this is quantified to give a numeric value called *fitness*. Those programs that do well are chosen to breed, and produce new programs for the new generation. The primary variation operators to perform transitions within the space of computer programs are crossover (e.g. subtree crossover) and mutation (e.g. point, bit-flip, subtree mutation) [5]. Like in other evolutionary algorithms, GP randomly generates individuals for the initial population. Two dominant methods are the full and grow, as well as the widely used combination of the two, known as Ramped half-and-half [5]. In both methods, the initial individuals are generated so that they do not exceed a user-specified maximum depth. The depth of a node is the number of edges that need to be traversed to reach the node starting from the tree's root node (the depth of the tree is the depth of its deepest leaf). Once a stopping criterion has been met the algorithm terminates and the best program is designated as the output of the run.

In some cases of prediction modelling, the trees produced by tree-generation algorithms are not comprehensible to users due to their *size* and *complexity* [6]. It is often desirable that the proposed approaches provide insight and understanding into the predictive structure of the data to be able to explain each individual prediction [20]. In [6], it is argued that the incomprehensibility of some models is caused by the model induction process being primarily based on *predictive accuracy* or *performance*. To address this concern, we use a multi-objective Genetic Programming algorithm to optimize decision trees for both classification performance and comprehensibility, without discriminating against either. MOO is a relatively new field in the area of telecommunications and it is difficult to apply an existing linear/single objective method to effectively tackle the Multiobjective Optimization Problem (MOP), giving a set of non-dominated solutions. The literature hosts several interesting approaches for tackling MOPs, with Multi-Objective Evolutionary Algorithms (MOEAs) [8] posing all the desired characteristics for obtaining a set of non-dominated solutions, in a single run. The two major classes of MOEAs are the Pareto-dominance based approaches [8] and the approaches based on decomposition [21]. Research studies that used GP approaches having MOEA characteristics for dealing with MOPs include the following: In [6], a Pareto-dominance based GP approach is used to optimize three objectives, i.e. classification accuracy, tree size and performance for medical data mining. [22] proposes a Pareto-dominance based GP variant, coined Traceless Genetic Programming for dealing with five multiobjective test problems. More recently in 2009, [23] have used a GP with MOEA based on Pareto dominance characteristics to automatically construct stochastic processes.

However, all research studies just mentioned use Pareto-dominance based approaches. Recently, a new and promising MOEA based on Decomposition (MOEA/D) [7] approach was proposed and it has shown a good performance in both continuous [7] and combinatorial problems [24], [25]. MOEA/D decomposes a MOP into a set of scalar subproblems and solves them using neighborhood information and scalar techniques, in a single run. In this paper, a GP with MOEA/D characteristics is proposed to find a good prediction model of vTEC over Cyprus, focusing in optimizing the performance (i.e. predictive accuracy) and complexity (i.e. comprehensibility measured in terms of tree size). To the best of our knowledge this is the first time that the vTEC prediction problem is studied in the context of MOO and a GP-MOEA/D based approach has never been applied to this problem before.

## III. PROBLEM DEFINITION AND MODEL

In this section, the characteristics of vertical Total Electron Content (vTEC) are introduced and particularly discussed for vTEC over Cyprus for a period of 11 years. The model parameters are also presented.

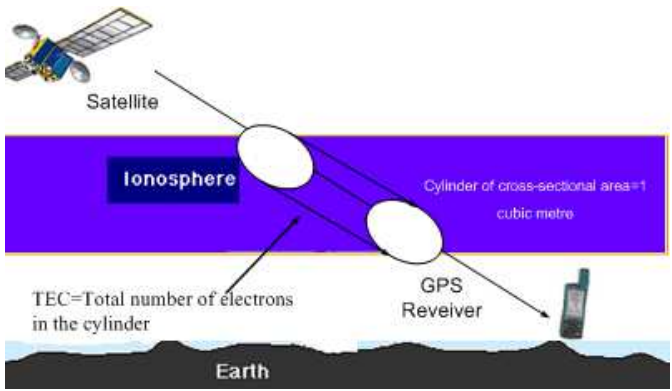

Fig. 1. Slant TEC representation.

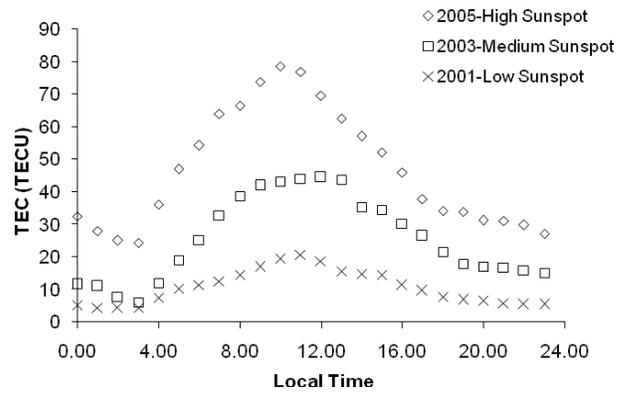

Fig. 3. Examples of diurnal variation of vTEC for low, medium and high solar activity.

*A. Total Electron Content Characteristics*

Dual-frequency GPS data recorded by GPS receivers enable an estimation of the Total Electron Content (TEC) measured in total electron content units, ($1\ TECU = 10^{16} electrons/m^2$). This is the total amount of electrons along a particular line of sight between the receiver and a GPS satellite in a column of $1 m^2$ cross-sectional area (illustrated in Figure 1) and represents a typical quantitative parameter of interest to GPS users. vTEC corresponds to the integral of the vertical electron density profile, an example of which is shown in Figure 2 from the ground to an infinite height (practically the height of the satellite). The analysis used in the present work to estimate vTEC from GPS data was carried out by means of the procedure developed by Ciraolo [26].

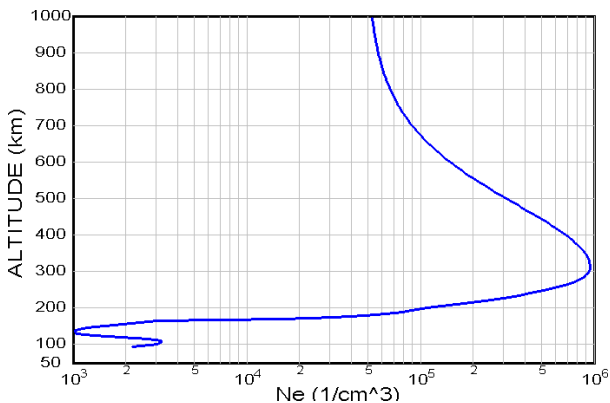

Fig. 2. Typical electron density profile of the ionosphere over Cyprus.

The electron density of free electrons within the ionosphere and therefore vTEC depend upon the strength of the solar ionizing radiation which is a function of time of day, season, geographical location and solar activity [1], [2]. Since solar activity has an impact on ionospheric dynamics which in turn influence the electron density of the ionosphere, vTEC also exhibits variability on daily, seasonal and long-term time scales in response to the effect of solar radiation. It is also subject to abrupt variations due to enhancements of geomagnetic activity following extreme manifestations of solar activity disturbing the ionosphere from minutes to days on a local or global scale.

The most profound solar effect on vTEC is reflected on its daily variation as shown in the typical examples for three days at different parts of the sunspot cycle in Figure 3. As it is clearly depicted in this figure, there is a strong dependency of vTEC on local time which follows a sharp increase of vTEC around sunrise and gradual decrease around sunset. This is attributed to the rapid increase in the production of electrons due to the photo-ionization process during the day and a more gradual decrease due to the recombination of ions and electrons during the night.

There is also a seasonal component in the variability of vTEC, which can be attributed to the seasonal change in extreme ultraviolet (EUV) radiation from the Sun. This can be clearly identified in Figure 4 for all daily noon values of vTEC collected for high and low solar activity periods (years 2001 and 2008). The long-term effect of solar activity on vTEC, which follows an eleven-year cycle, is also clearly shown in both Figures 3 and 4, in which we can observe higher vTEC variability for higher solar activity in both diurnal and seasonal time-scales.

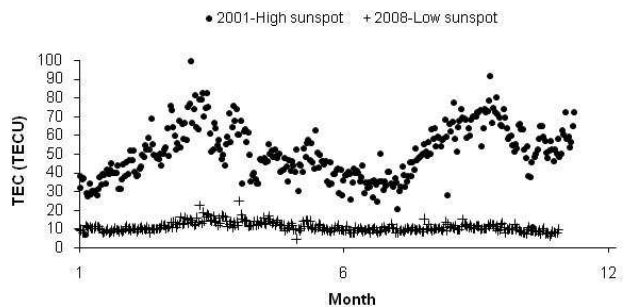

Fig. 4. Seasonal variation of all vTEC values at 12:00.

*B. Model Parameters*

The diurnal variation of vTEC is clearly evident by observing Figure 3. We therefore include hour number as an input to the model. The hour number, $hour$, is an integer in the range $0 \leq hour \leq 23$. In order to avoid unrealistic discontinuity at the midnight boundary, $hour$ is converted into its quadrature components according to:

$$sinhour = sin(2\pi \frac{hour}{24}) \quad (1)$$

and

$$coshour = cos(2\pi \frac{hour}{24}) \quad (2)$$

A seasonal variation is also an underlying characteristic of vTEC as shown in Figure 4 and is represented by day number $daynum$ in the range $1 \leq daynum \leq 365$. Again to avoid unrealistic discontinuity between December $31^{st}$ and January $1^{st}$, $daynum$ is converted into its quadrature components according to:

$$sinday = sin(2\pi \frac{daynum}{365}) \quad (3)$$

and

$$cosday = cos(2\pi \frac{daynum}{365}) \quad (4)$$

Long-term solar activity has a prominent effect on vTEC. To include this effect in the model specification we need to incorporate an index, which represents a good indicator of solar activity. In ionospheric work the 12-month smoothed *sunspot number* is usually used, yet this has the disadvantage that the most recent value available corresponds to vTEC measurements made six months ago. To enable vTEC data to be modeled as soon as they are measured, and for future predictions of vTEC to be made, the monthly mean sunspot number values were modeled using a smooth curve defined by a summation of sinusoids.

## IV. GENETIC PROGRAMMING + MOEA/D

In this section the problem representation is introduced and the vTEC prediction problem is formulated in the context of MOO. The description of the evolutionary algorithm employed, coined GP-MOEA/D, follows. GP-MOEA/D is a standard elitist (i.e. the best is always preserved), generational (i.e. populations are arranged in generations, not steady-state), panmictic (i.e. no mating restrictions) [27] Multi-Objective Evolutionary Algorithm based on Decomposition (MOEA/D) characteristics. Interested readers are referred to [7] for details on MOEA/D.

### A. Problem Representation and MOO Formulation

In this paper, a prediction model is represented by Ramped-half-and-half trees $X$ with an initial maximum depth of 6 that are allowed to grow up to depth of 12 during evolution. The models are evolved into a constrained, problem-specific user-defined language. The variables and constants of the model are leaves in the tree (collectively named as terminal set $T$), whilst arithmetic operators are internal nodes (collectively named as function set $F$). It is common in the GP literature to represent expressions in the prefix notation similar to that used in LISP or Scheme. For example, x+3*y becomes (+ x (* 3 y))). This representation eases the expression-tree data structure formation, and its manipulation during the application of variation operators, which will be explained soon. GP finds out how well a program works by running it, and then comparing its behaviour to some ideal, i.e. exact measurements.

In this paper, we are interested in how well a model $X$ predicts vTEC in a given data set $D$ of size $n$, denoted as $X(in_j) : j = 1,\ldots,n$, where $in_j$ is the vector consisting the input parameters (defined in Subsection III-B) of instance $j$ in $D$. This comparison is quantified to give a numeric fitness value of tree $X$, which in our case is the $RMSE(X,D)$. Besides, on the one hand, it is accepted that smaller decision trees are more comprehensible and have better generalization capabilities to adapt to the variations of the parameters in the whole data set. On the other hand, the bigger the tree size is the less generalized (and more complex) the tree is, and consequently the more biassed in terms of RMSE (i.e. more accurate prediction structures). Therefore, the $RMSE(X,D)$ and the size of the tree, i.e. $Size(X)$ are conflicting objectives and should be optimized in the context of MOO. The proposed vTEC prediction MOP formulation is as follows:

**Given:**
- $D$: data set
- $T$: terminal set
- $F$: function set

**Decision variables of a prediction tree $X$:**
- variables and constants from terminal set $T$
- operands from function set $F$
- the connections between variables/constants and operands.

**Objectives**: Minimize RMSE and the size of tree $X$:

$$\min \ RMSE(X,D) = \sqrt{\frac{\sum_{j=1}^{n}(X(in_j) - vTEC_j)^2}{n}} \quad (5)$$

where $in_j$ is the vector consisting the input parameters of instance $j$ in data set $D$ and $vTEC_j$ is the corresponding measured vTEC value.

$$\min \ Size(X) = |X| \quad (6)$$

which is the number of nodes composing the tree solution $X$.

In a MOP [8], there is no single solution $X$ that optimizes all objectives simultaneously, but a set of trade-off candidates. The set of trade-off solutions is often defined in terms of Pareto Optimality [8]. That is, considering a minimization MOP[1] with $m$ decision variables and $n$ objectives:

- **Definition 1** (Pareto dominance). For any two decision variable vectors $x = (x_1,\ldots,x_m)^T$ and $y = (y_1,\ldots,y_m)^T$, $x$ is said to *dominate* $y$, denoted by $x \prec y$, if and only if $f_i(x) \leq f_i(y)$ for every $i \in \{1,2,\ldots,n\}$ and $f_j(x) < f_j(y)$ for at least one index $j \in \{1,2,\ldots,n\}$. $x$ is said to be *nondominated*, if there is no $y \in \Omega$ which dominates $x$, where $\Omega$ is the objective space.

- **Definition 2** (Pareto optimality). An objective vector $u = (u_1,\ldots,u_n)^T$ is said to be *(globally) Pareto-optimal* if there does not exist another objective vector $v = (v_1,\ldots,v_n)^T$ such that $v$ dominates $u$, the latter is then called the *Pareto objective vector*. The set of all Pareto-optimal objective vectors is called the *Pareto-optimal front*, denoted by PF. The set of all Pareto-optimal solutions in the decision space is called the *Pareto-optimal set*, denoted by PS.

---
[1] the Pareto Optimality for MOPs with maximization objectives can be defined similarly.

*B. The Proposed Methodology*

The proposed GP-MOEA/D proceeds as in Algorithm 1 and is described in the following.

*1) Setup-Decomposition:* Initially, the MOP should be decomposed into $m$ subproblems by adopting any technique for aggregating functions [7], e.g. the Tchebycheff approach used here. In this paper, the $i^{th}$ subproblem is in the form

$$\text{maximize} \quad g^i(X|w_j^i, z^*) = max\{w_j^i|f_j(X) - z_j^*|\} \quad (7)$$

where $f_j$, $j = 1, 2$ are the objectives of the MOP in Subsection IV-A, $z^* = (z_1^*, z_2^*)$ is the reference point, i.e. the maximum objective value $z_j^* = max\{f_j(X) \in \Omega\}$ of each objective $f_j, j = 1, 2$ and $\Omega$ is the decision space. For each Pareto-optimal solution $X^*$ there exists a weight vector $w$ such that $X^*$ is the optimal solution of (7) and each solution is a Pareto-optimal solution of the MOP in Subsection IV-A. For the remainder of this paper, we consider a uniform spread of the weights $w_j^i$, which remain fixed for each subproblem $i$ for the whole evolution and $\sum_{j=1}^{2} w_j^i = 1$. By decomposing the MOP into a set of scalar subproblems one can predict the objective preference of a particular prediction tree $X$ and therefore its position in the objective space, considering the $w^i$ weight coefficient of a subproblem $i$. For example, the $g^i(X|w_j^i, z^*)$ with $w_j^i = (1, 0)$ means that the subproblem $g^i$ focuses in optimizing objective function $f_1$ (in this case RMSE), ignoring the other objective function and consequently utilizing all its effort in obtaining a prediction tree of minimum RMSE. In the same way, $g^i(X|w_j^i, z^*)$ with $w_j^i = (0, 1)$ focuses in prediction trees of just minimum size. The goal, however, in vTEC prediction problem is to obtain the solutions of these extreme cases as well as the trade-off between them, e.g. $w_j^i = (0.3, 0.7)$. Consequently, appropriate scalar strategies can be employed and controlled to optimize different feasible areas of the objective space accordingly. Note that, this beneficial procedure cannot be utilized by any non-decompositional MOEA framework.

*2) Setup-10 fold validation:* The data-set was segmented in 10 continuous folds similarly to [10]. In each cross-validation cycle, 9 folds were used as the training set, whereas the evolved model was tested on the remaining $10^{th}$ fold. The training set was further randomly divided into two data-sets (with no overlapping): the fitness evaluation data-set, with 67% of the training data, and the validation data-set with the remaining 33%.

*3) Initialization:* In Step 1 of Algorithm 1, we adopt a random method to generate $m$ solutions for the initial internal population (i.e $IP_0$). Namely, a tree solution $X$ is initiated by using a Ramped-half-and-half tree creation with a maximum depth of 6 to perform a random sampling of rules. Each tree $X$ is composed of variables and constants from terminal set $T$ as well as operands from function set $F$. Each tree solution $X \in IP_0$ is then evaluated using the training set generated during setup.

---

**Algorithm 1** The GP+MOEA/D

**Input:**
- vTEC parameters, terminal set $T$;
- GP primitive language, function set $F$;
- $m$ : population size and number of subproblems;
- weight vectors $(w_j^1, ..., w_j^m)$, $j = 1, 2, 3$;
- the maximum number of generations, $gen_{max}$;

**Output:** an optimal prediction model $X^*$.

**Step 0-Setup:**
- Decompose the MOP;
- Generate the 10-fold validation sets;
- Set $EP := \emptyset$; $gen := 0$; $IP_{gen} := \emptyset$;

**Step 1-Initialization:** Uniformly randomly generate an initial set of prediction trees $IP_0 = \{X^1, \cdots, X^m\}$, known as initial internal population, by using $T$ and $F$;

**Step 2:** For $i = 1, \ldots m$ **do**

    **Step 2.1-Genetic Operators:** Generate a new solution (i.e. prediction tree) $Y$ using the genetic operators.
    **Step 2.2-Evaluation on Training Set:** Evaluate $Y$ using the training set.
    **Step 2.3-Update Populations:** Use $Y$ to update $IP_{gen}$, $EP$ and the $T$ closest neighbor solutions of $Y$.

**Step 3-Stopping criterion:** If stopping criterion is satisfied, i.e. $gen = gen_{max}$, **then** stop and forward $EP$, **otherwise** $gen = gen + 1$, go to Step 2.

**Step 4-Evaluation on Validation Set:** Evaluate all solutions $Z \in EP$ using the validation set.

**Step 5-Output:** Evaluate solution $X^* \in EP$ having the lowest RMSE with respect the validation set and evaluate it using the $10^{th}$ fold.

---

*4) Genetic Operators:* In Step 2.1 of Algorithm 1, the genetic operators are then invoked on $IP$ for offspring reproduction for each subproblem $g^i$, where $i = 1$ to $m$. Initially, the popular tournament selection [4] is utilized. Tournament selection randomly chooses a finite size set of tree solutions $X$ from the current population $IP_{gen}$. From this set the solution with the best fitness, i.e. $g^i(X|w_j^i, z^*)$, is selected for reproduction and forwarded to the breeding operators. In this paper, neither recombination, nor reproduction was used for breeding, but just mutation. Particularly, a mixture of mutation-based variation operators is employed, where subtree mutation is combined with point-mutation, for generating a new solution $Y$. The two mutation operators are probabilistically selected using a pre-defined parameter.

*5) Evaluation (training set) and update of populations:* In Step 2.2, the new solution $Y$ is evaluated using the training set generated in the setup phase. Then the update of populations, which is processed in two steps, follows. (1) Update $IP$, which keeps the best solution found so far for each subproblem $i$, $IP/\{X^i\}$ and $IP \cup \{Y^i\}$ if $g_i(Y^i|w^i, z^*) < g^i(X^i|w^i, z^*)$, otherwise $X^i$ remains in $IP$. (2) Update the External Population ($EP$), which stores all the non-dominated solutions found so far during the search. $EP = EP \cup \{Y^i\}$ if $Y^i$ is not dominated by any solution $X^j \in EP$ and $EP = EP/\{X^j\}$, for all $X^j$ dominated by $Y^i$. The two-objective sort conducted in this step is in order to extract a set of non-dominated individuals [8] (Pareto Front), with regards to the lowest fitness evaluation data-set RMSE, as well as the smallest model complexity in terms of expression- tree size (measured by the number of tree-nodes). The rationale behind this is to create selection pressure towards accurate but simpler prediction models that have the potential to generalise better. These non-dominated individuals are then evaluated on the validation data-set, with the best-of-generation prediction model selected as the one of these with the smallest RMSE. During tournament selection based on the fitness evaluation data-set performance, we used the model complexity as a second point of comparison in cases of identical error rates.

*6) Stopping criterion, evaluation (validation set) and output:* In Step 3, the search stops after a pre-defined number of generations, $gen_{max}$. When the termination criterion in Step 3 is satisfied, the $EP$, which holds all non-dominated solutions found during the search is evaluated using the validation set (generated in setup) in Step 4. Finally, in Step 5, the best solution $X^*$ found in terms of RMSE, evaluated using the validation set, is evaluated in the $10^{th}$ fold (generated during setup) and output as the best prediction model.

## V. Experimental Methodology

### A. Data Set

The vTEC data-set used in this work consist of around 60000 values recorded between 1998 and 2009. In this paper, the data-set was segmented in 10 continuous folds similarly to [10]. In each cross-validation cycle, 9 folds are used as the training set, whereas the evolved model is tested on the remaining $10^{th}$ fold. The training set is further randomly divided into two data-sets (with no overlapping): the fitness evaluation data-set, with 67% of the training data, and the validation data-set with the remaining 33%. The fitness measure (of Step 2.2) consists of minimising the RMSE on the fitness evaluation data-set.

### B. Algorithms

Two GP-based approaches are used for evaluating the performance of our GP+MOEA/D based approach:

(i) The conventional single objective GP (i.e. sGP) that uses all the GP characteristics of the proposed approach described in Section IV except the multi-objective optimization characteristics of Algorithm 1, i.e. Steps 2.3 and 4 related to Multi-objective Pareto-dominance ranking. Particularly, this approach evolves a prediction model in the training set and validates it on the validation set. The evolution stops when no further convergence is noticed in five consecutive generations.

(ii) The Pareto-dominance based GP, i.e. GP-NSGAII which is a GP approach having the characteristics of the state-of-the-art in MOEAs based on Pareto dominance, the Non-Dominated Sorting Genetic Algorithm II (NSGA-II) [9]. Particularly, NSGA-II maintains a population $IP_{gen}$ of size $m$ at each generation $gen$, for $gen_{max}$ generations. NSGA-II adopts the same evolutionary operators (i.e. selection, crossover and mutation) for offspring reproduction as MOEA/D. The key characteristic of NSGA-II is that it uses a fast non-dominated sorting and a crowded distance estimation for comparing the quality of different solutions during selection and to update the $IP_{gen}$ and the $EP$. We refer interested readers to [9] for details.

The GP-based algorithms use tournament selection with a tournament size of 7. Evolution proceeds for 50 generations, and the population size is set to 1000 individuals. Ramped-half-and-half tree creation with a maximum depth of 6 is used to perform a random sampling of rules during run initialisation. Throughout evolution, expression-trees are allowed to grow up to depth of 12. The evolutionary search employs a mixture of subtree mutation combined with point-mutation; with the probability governing the application of each set to 0.6 in favour of sub-tree mutation. The primitive language consisted of the basic arithmetic operators (+, -, *, /) serving as the function set, whereas the terminal set consisted of the five independent variables described in Section III.

Finally, the performance of the proposed approach is studied against the previously proposed Neural Network (NN) approach [10]. The NN approach has a fully connected two-layer structure, with 5 input, 10 hidden and 1 output neurons. Both the hidden and output neurons of the NN consisted of hyperbolic tangent sigmoid activation functions. The number of hidden neurons was determined by trial and error. The training algorithm used was the Levenberg-Marquardt back propagation algorithm.

All approaches were coded in Java and run on an Intel Pentium 4 3.2 GHz Windows XP server with 1.5 GB RAM. We performed 50 independent evolutionary runs for each test fold, in order to account the stochastic nature of the adaptive search algorithms, and obtain statistically meaningful results.

### C. Performance Metrics

For evaluating the performance of the approaches the RMSE metric is mainly utilized, as well as some statistical metrics, e.g. mean, max, min and standard deviation. Furthermore, the Multi-Objective Evolutionary Algorithms (i.e. MOEA/D and NSGA-II) were studied according to the quality and diversity of their PF obtained during evolution. Since MOEAs generate a set of solutions for approximating the PF, it is not easy to compare the algorithms performances and there is not a single metric that can satisfy all requirements [9], [28], [29]. For this purpose, the following three metrics are adopted:

The $\Delta$-metric [9] measures the extent of spread achieved among the obtained solutions. In the case of two objectives, the $\Delta$ value of a set of candidate solutions $A$ is defined as follows:

$$\Delta(A) = \frac{d_f + d_l + \sum |d_j - \overline{d}|}{d_f + d_l + |A|\overline{d}},$$

where $d_f$ and $d_l$ are the extreme Pareto optimal solutions in the objective space, $d_j$ is the distance between two neighboring solutions and $\overline{d}$ is the mean of all the distribution. The smaller the $\Delta(A)$ metric is, the better the diversity performance of $A$. $\Delta(A)$=0 means a uniform spread of solutions in the objective space.

A straightforward comparison metric between two sets of non-dominated solutions $A$ and $B$ is the C-metric [9], [29]. The $C(A, B)$ metric, which is usually considered as a MOEA quality metric, evaluates the ratio of the non-dominated solutions in $A$ dominated by the non-dominated solutions in $B$,

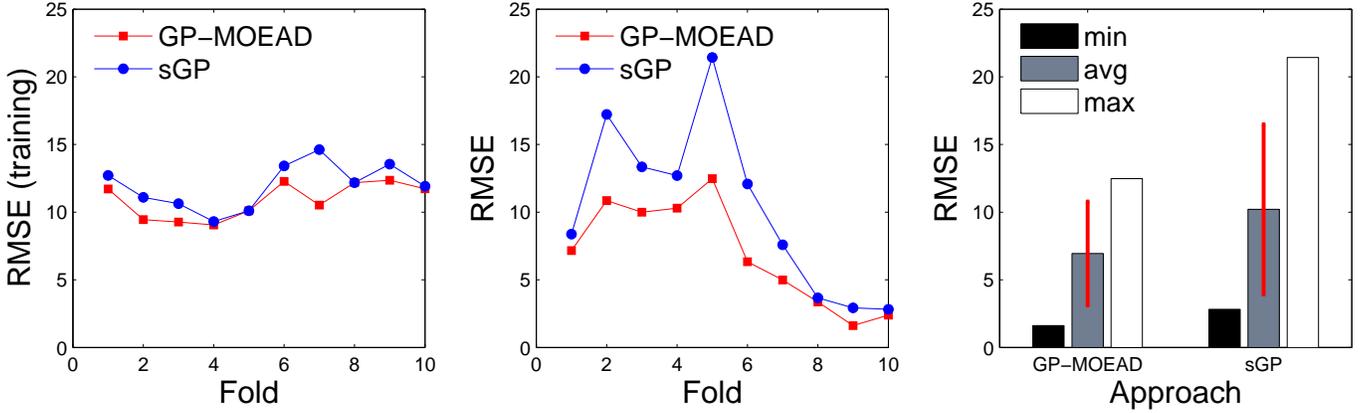

Fig. 5. Conventional Single-objective GP versus GP-MOEA/D. (left) RMSE on the training set per fold. (center) The final RMSE of the proposed models of each approach per fold. (right) The min, max, average and standard deviation of RMSE on all 10 folds.

divided by the total number of nondominated solutions in $A$. Hence,

$$C(A, B) = \frac{|A - \{x \in A | \exists y \in B : y \prec x\}|}{|A|}.$$

The smallest $C(A, B)$ is, the better the $A$. Note that $C(A, B) \neq 1 - C(B, A)$.

Another commonly used metric, usually considered in cases of real-life discrete optimization problems [30], [31], is the number of Non-Dominated Solutions ($NDS(A)$) in set A, i.e.

$$NDS(A) = |A|.$$

In the type of problem considered in this paper it is very difficult to obtain many different $NDS$s. Therefore, a high number of $NDS(A)$ is desirable to provide an adequate number of Pareto optimal choices. However, the $NDS$ should be considered in combination with other metrics (e.g. $\Delta$ and $C$ metrics), since it is usually desirable to have a high number of $NDS$ when the solutions is of high quality and spread in the objective space. In contrast, and usually in cases of continuous optimization [7], a high number of $NDS$ is not desirable, since the decision making procedure becomes more complicated.

## VI. Experimental Results and Discussion

The primary goal of our experimental studies is to investigate the performance of our GP-based approach in designing a prediction model for vTEC over Cyprus with which to approximate the measured values, compared to other GP-variants and the previously proposed Neural Network based model.

### A. Conventional single-objective GP versus GP-MOEA/D

Initially, the proposed GP-MOEA/D is compared with the conventional single-objective GP (described in Subsection V-B). Figure 5 shows the performance of the two approaches during evolution using the training set (left), based on the RMSE of the final proposed prediction model on each test fold (center) and based on the minimum, maximum and average RMSE of the proposed models on all 10 folds (right). The results clearly demonstrate the superiority of the proposed GP-MOEA/D due to its ability in increasing the selection pressure towards prediction models that have the potential to generalise better. The left subfigure of Figure 5 show that the two approaches provide similar RMSE during evolution and when evaluated on the training set. However, the proposed prediction models of GP-MOEA/D (in the center subfigure) perform better than those of the sGP on the final RMSE on each fold. The right subfigure supports these observations, since GP-MOEA/D provides a lowest minimum, maximum and mean RMSE considering all folds, having a smallest standard deviation as well.

### B. GP-NSGA-II versus GP-MOEA/D

In this subsection, we have evaluated the performance of the proposed GP-MOEA/D (i.e. GP with the decompositional approach MOEA/D) against the GP-NSGAII (i.e. GP with the Pareto-dominance based approach NSGA-II described in Subsection V-B.) The two MOEA approaches, using the training set, have obtained a set of non-dominated prediction models, i.e. the PF, for each fold as illustrated in Figure 6.

Figure 6 shows the Pareto-optimal solutions of each approach per fold, where the solutions of GP-MOEA/D are denoted by red crosses and those of GP-NSGAII with green diamonds. The results show that GP-MOEA/D's PF outperforms the PF of GP-NSGA-II in most cases. The solutions of the proposed approach are of better quality as well as diversity, providing a higher number of non-dominated prediction models that are spread in the objective space indicating a better exploration. In most cases, the two approaches perform similarly for high RMSE and low model sizes. However, the decompositional nature of GP-MOEA/D forces the proposed approach to converge towards complex models of lower RMSE more efficiently than GP-NSGAII, giving more prediction model choices. The observations just mentioned are also supported by the statistical results summarized in Table I, where the best results are denoted in bold.

The statistical results show that GP-MOEA/D's Pareto-optimal solutions dominate all solutions obtained by GP-

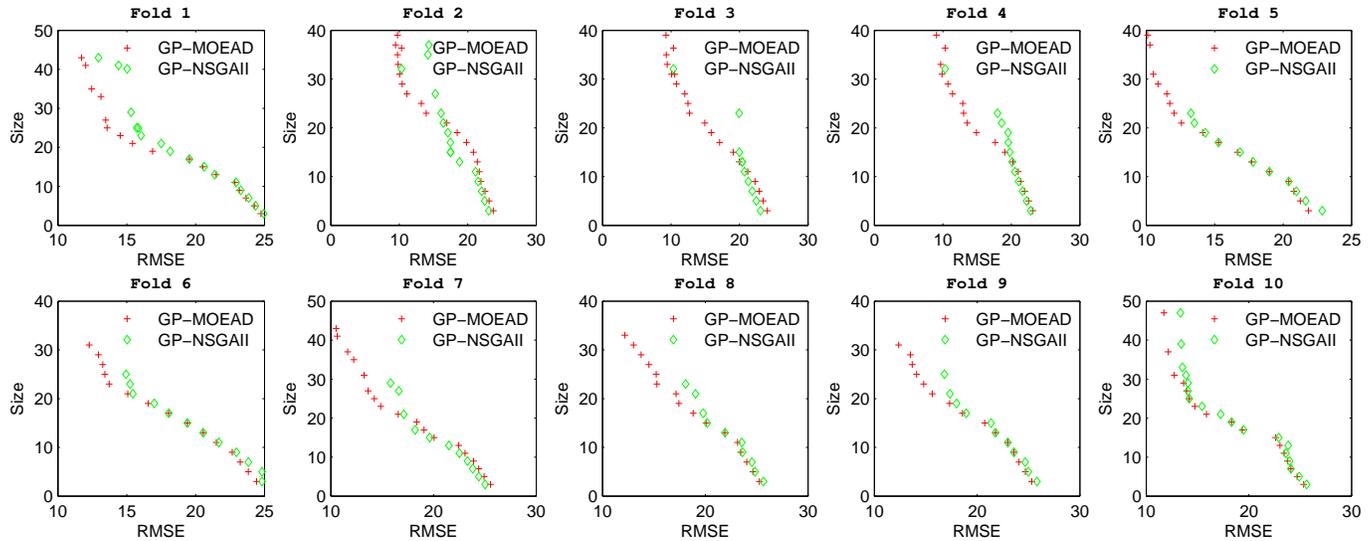

Fig. 6. GP-MOEA/D versus GP-NSGA-II, a per fold comparison with respect to the final PF obtained by each Multi-objective Optimization GP approach.

TABLE I
GP-MOEA/D (M) vs. GP-NSGA-II (N) on all 10 folds (Fold1-10)

| Metric: | C(N,M) | C(M,N) | $\Delta$(N) | $\Delta$(M) | NDS(N) | NDS(M)) |
|---|---|---|---|---|---|---|
| Fold1: | 0.0625 | **0.8824** | 18.3674 | **0.6737** | 16.0 | **17.0** |
| Fold2: | **0.7333** | 0.4211 | 3.4282 | **0.4664** | 15.0 | **19.0** |
| Fold3: | **0.5556** | 0.3158 | 3.63 | **0.5459** | 9.0 | **19.0** |
| Fold4: | 0.5455 | **0.2941** | 3.633 | **0.4393** | 11.0 | **17.0** |
| Fold5: | 0 | **0.4706** | 3.6582 | **0.5829** | 11.0 | **17.0** |
| Fold6: | 0 | **0.6** | 3.6369 | **0.4428** | 12.0 | **15.0** |
| Fold7: | **0.7273** | 0.2222 | 1.8097 | **0.6552** | 11.0 | **18.0** |
| Fold8: | 0 | **0.6875** | 1.5351 | **0.511** | 10.0 | **16.0** |
| Fold9: | 0 | **0.6** | 1.8994 | **0.6013** | 11.0 | **15.0** |
| Fold10: | 0 | **0.8235** | 3.6815 | **0.6470** | 18.0 | **17.0** |
| mean: | 0.2624 | **0.5317** | 4.5279 | **0.5566** | 12.4 | **17.0** |
| std: | 0.3314 | **0.2252** | 4.9427 | **0.0890** | 2.9136 | **1.4142** |

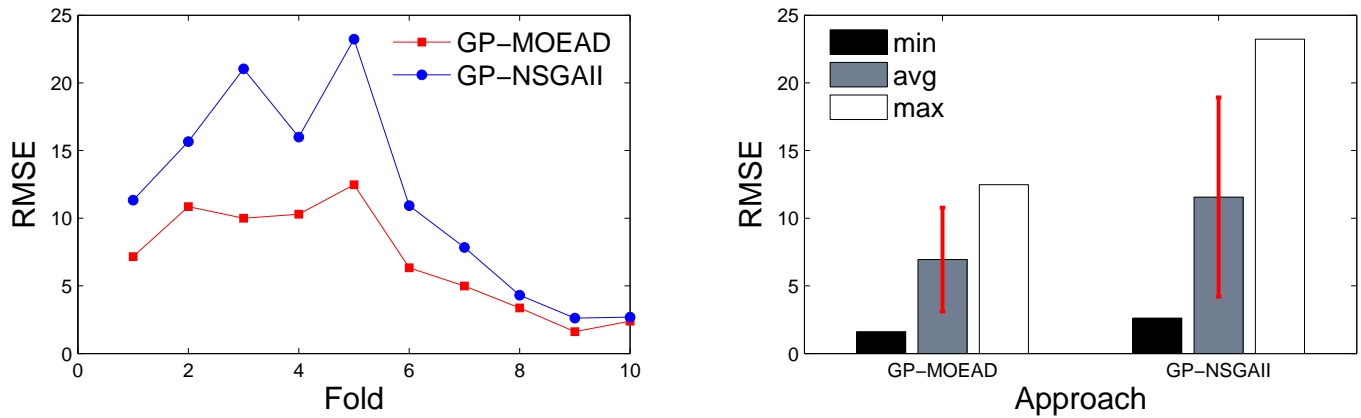

Fig. 7. GP-NSGAII versus GP-MOEA/D. (left) The final RMSE of the proposed prediction models of each approach per fold. (right) The min, max, average and standard deviation of RMSE on all 10 folds.

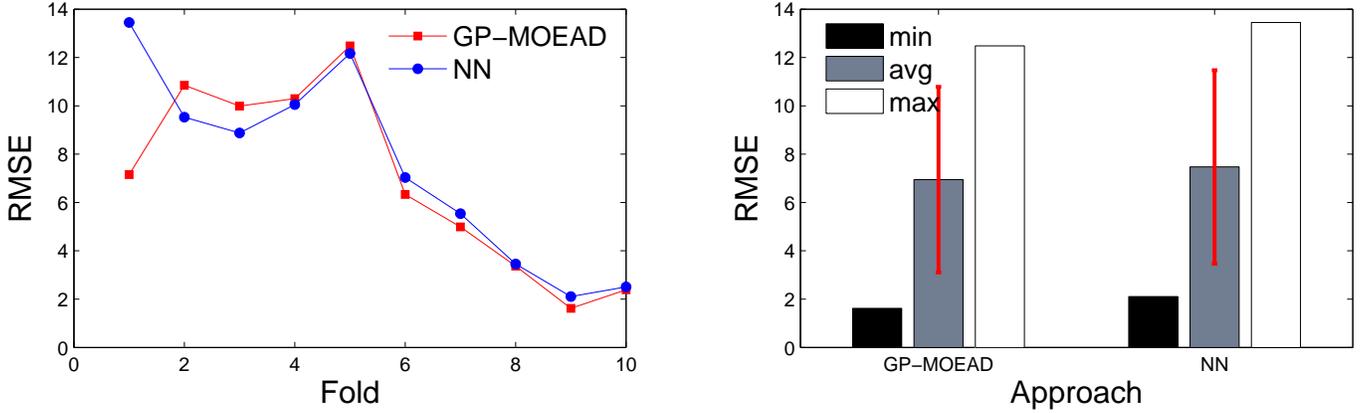

Fig. 8. Neural Network (NN) versus GP-MOEA/D. (left) The final RMSE of the proposed prediction models of each approach per fold. (right) The min, max, average and standard deviation of RMSE on all 10 folds.

NSGA-II in four out of ten folds, providing better quality in two more (this is indicated by the C-metric in columns two and three of Table I). On average, the Pareto-optimal solutions of GP-MOEA/D dominate 53% of the Pareto-optimal solutions obtained by GP-NSGA-II having a lower standard deviation as well. In terms of diversity the superiority of GP-MOEA/D is clearer since it provides a more diverse PF on all 10 folds (this is indicated by the D-metric in columns four and five of Table I), giving a higher number of non-dominated solutions (the NDS-metric in columns six and seven of Table I) and consequently more prediction models choices. The PF obtained by GP-MOEA/D is about nine times more diverse with five Pareto-optimal solutions more than the PF obtained by GP-NSGA-II, on average.

Finally, Figure 7 shows a comparison of the two MOEAs with respect to the RMSE of the final proposed prediction model on each fold (left) and based on the minimum, maximum and average RMSE of the proposed models on all 10 folds (right). The results show that GP-MOEA/D obtains a better prediction model on all ten folds. GP-MOEA/D provides around 50% lower RMSE compared to the one of GP-NSGAII in the worst case (i.e. the maximum RMSE obtained by both MOEAs is in fold 5), around 20% lower RMSE in the best case (i.e. the minimum RMSE obtained by both MOEAs is in fold 9) and about 37% lower RMSE, on average.

### C. Neural Networks versus GP-MOEA/D

Based on the conclusions drawn in Subsections VI-A and VI-B, one can say that the best GP with MOEA characteristics approach presented in this paper is the GP-MOEA/D. In this subsection, the GP-MOEA/D is compared with a Neural Network based approach, which was already shown to be efficient in predicting vTEC over Cyprus in [10]. The comparison between the two approaches is illustrated in Figure 8 with respect to the RMSE of the final proposed prediction model on each fold (left) and based on the minimum, maximum and average RMSE of the proposed models on all 10 folds (right). The results show that GP-MOEA/D performs better than the Neural Network approach in six out of ten folds.

GP-MOEA/D provides around 7.5% lower RMSE compared to the one of Neural Network in the worst case (i.e. the maximum RMSE obtained by GP-MOEA/D is in fold 5 and the maximum RMSE obtained by NN is in fold 1), around 24% lower RMSE in the best case (i.e. the minimum RMSE obtained by both approaches is in fold 9) and about 7% lower RMSE, on average.

Additionally, it is important to note that all approaches converge towards similar values in the last three folds of all experimental studies (i.e. Figures 5, 7 and 8). This is due to the fact that the variability of the vTEC on these three folds is low, and therefore it is much easier to obtain more accurate predictions.

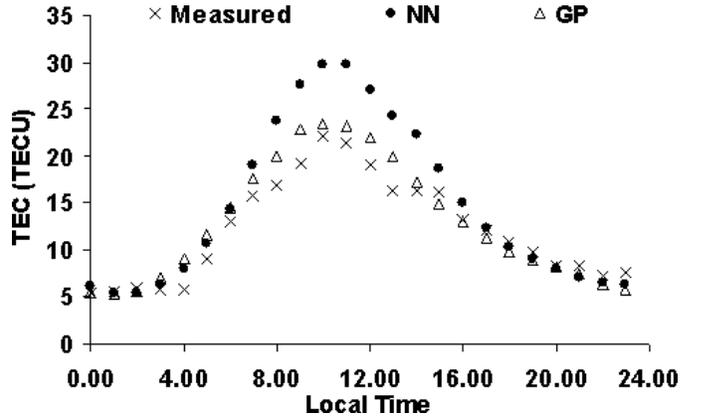

Fig. 9. Measured vs. GP and NN predicted values of diurnal variation of vTEC - Case 1.

### D. Measured (exact) versus GP-predicted values

Finally, in this subsection we demonstrate the effectiveness and efficiency of GP-MOEA/D in approximating the actual measurements of the diurnal variation of vTEC over Cyprus with respect to the Neural Network approach.

Figures 9 and 10 show the good performance of GP-MOEA/D in approximating vTEC during a period of 24 hours in different days of the year. The results support the observations of Subsection VI-D that GP-MOEA/D performs better than Neural Network approach in most cases. GP-MOEA/D approximates the measured values of vTEC by around 2% in Case 1 of Figure 9 and by 4% in Case 2 of Figure 10, where the Neural Network approach approximates the measured vTEC values of Cases 1 and 2 by 4% and 10%, respectively. From the ionospheric perspective, in Case 1, we observe that during the night both models exhibit similar performance but during the day where the variability in the ionosphere is significantly higher, GP-MOEA/D clearly outperforms the Neural Network approach. This is also true for Case 2 in addition to the fact that GP-MOEA/D significantly outperforms the Neural Network approach also after sunset.

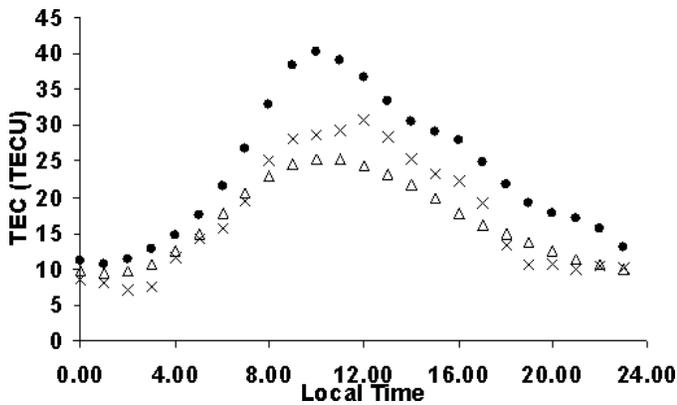

Fig. 10. Measured vs. GP and NN predicted values of diurnal variation of vTEC - Case 2.

## VII. CONCLUSIONS

In this paper, a Genetic Programming (GP) based approach is used to design a prediction model for Total Electron Content over Cyprus in the context of Multi-Objective Optimization. Particularly, a panmictic, generational, elitist GP with an expression-tree representation, having the characteristics of the Multi-Objective Evolutionary Algorithm based on Decomposition (MOEA/D), coined GP-MOEA/D is used. A prediction model is developed based on a data set obtained during a period of eleven years covering a full sunspot cycle. The experimental results have shown the superiority of the proposed approach with respect to a conventional (Single Objective Optimization) GP approach, a GP having the characteristics of the Pareto-dominance NSGA-II approach and a Neural Network (NN) approach. The GP-model has shown a good approximation of the different time-scales in the variability of the modelled parameter and it has outperformed its counterparts.

There are a number of avenues for future research. For example, it will be interesting to investigate different genetic operators and primitive languages to further improve the performance of the GP approach. Moreover, the hybridization of the GP with NNs and the design of a more robust approach is also a future possibility.